\documentclass[10pt,twocolumn,letterpaper]{article}

\usepackage{iccv}
\usepackage{times}
\usepackage{epsfig}
\usepackage{graphicx}
\usepackage{amsmath}
\usepackage{amssymb}

\usepackage{multirow}
\usepackage{amsmath,amssymb}

\usepackage[breaklinks=true,bookmarks=false]{hyperref}

\iccvfinalcopy 

\def\iccvPaperID{****} 

\begin{document}

\title{Remote Health Coaching System and Human Motion Data Analysis for Physical Therapy with Microsoft Kinect}

\author{Qifei Wang \& Gregorij Kurillo\\
University of California, Berkeley\\
Berkeley, CA 94720, USA\\
{\tt\small \{qifei.wang, gregorij\}@eecs.berkeley.edu}
\and
Ferda Ofli\\
Qatar Computing Research Institute\\
Doha, Qatar\\
{\tt\small fofli@qf.org.qa}
\and
Ruzena Bajcsy\\
University of California, Berkeley\\
Berkeley, CA 94720, USA\\
{\tt\small bajcsy@eecs.berkeley.edu}
\thanks{This research was supported by the National Science Foundation (NSF) under Grant No. 1111965.}
}

\maketitle

\begin{abstract}
   This paper summarizes the recent progress we have made for the computer vision technologies in physical therapy with the accessible and affordable devices. We first introduce the remote health coaching system we build with Microsoft Kinect. Since the motion data captured by Kinect is noisy, we investigate the data accuracy of Kinect with respect to the high accuracy motion capture system. We also propose an outlier data removal algorithm based on the data distribution. In order to generate the kinematic parameter from the noisy data captured by Kinect, we propose a kinematic filtering algorithm based on Unscented Kalman Filter and the kinematic model of human skeleton. The proposed algorithm can obtain smooth kinematic parameter with reduced noise compared to the kinematic parameter generated from the raw motion data from Kinect.
\end{abstract}

\section{Introduction}

In the last decade, we witness increasing interests for remote monitoring technologies in health care. Remote monitoring captures the activity and other anthropometric data from the subjects. It can perform online analysis of these data, provide health-care feedback to the subjects monitored, and also represent the data to the doctor for further analysis. One typical application is in the physical therapy, especially for the elderly subjects, the doctors need to monitor the subjects’ physical performance at home during the training sessions and provide further treatments based on the subjects’ performance.

Human performance capture systems have been widely studied and applied in many applications. However, traditional motion capture systems require particular system setup and markers attached to the subjects monitored which make it impractical for recording motion data at home. Other motion sensors, like the inertia sensors are sensitive to the noise make them also not suitable for motion capture in human daily activities. The emerging accessible and affordable sensing technologies, e.g. the depth sensor, such as Microsoft Kinect, make it possible to capture the human’s motion data at home \cite{Hondori2014}.

Recently, we built an interactive exercise coaching system based on Microsoft Kinect \cite{Ofli2015} intended to encourage elderly adults to perform physical therapy at home. Since this type of sensors were not designed for medical purpose, the acquired motion data are usually noisy with low fidelity for the motion analysis. For example, the joint position data from Microsoft Kinect usually exhibits significant jitters caused by low depth accuracy, occlusions, ambiguity, and loss of tracking. The body segment lengths are also vary during the motion. In order to perform online analysis of subjects’ performance and feedback to the subjects, we further studied the motion data accuracy from Microsoft Kinect. We also studied the kinematic parameter extraction from the motion data captured by Kinect.

This paper is organized as following: Section 2 introduces the interactive system we have built for the remote physical therapy; Section 3 introduces the work we have done on motion data accuracy evaluation of Microsoft Kinect; Section 4 describes our work on the human kinematic parameters extraction from Kinect motion data; Conclusions are summarized in Section 5.


\section{Interactive Exercise Coaching System}
\begin{figure}[t]
	\begin{center}
		\includegraphics[width=0.9\linewidth]{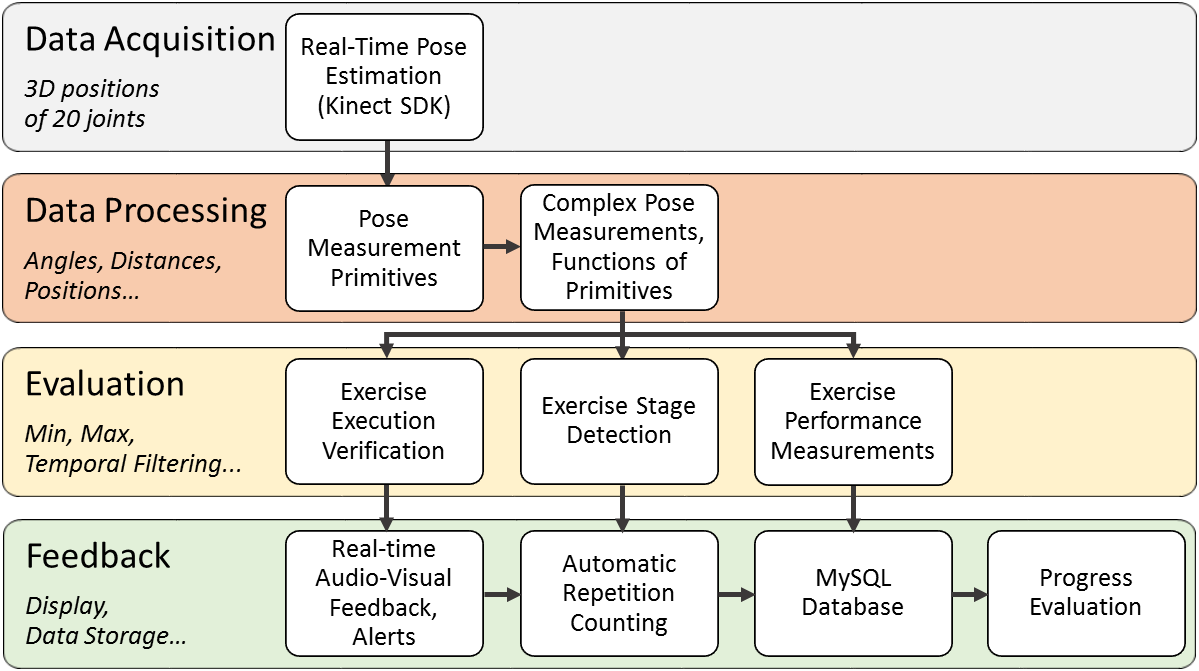}
	\end{center}
	\caption{Architecture of our interactive health coaching system.}
	\label{fig:fig1_system}
\end{figure}

This system is built for automated exercise coaching of elderly users that would motivate them and track their physical performance when they perform physical therapy at home. The system contain four main modules: data acquisition, data processing, performance evaluation, and feedback. The pipeline of this system is shown in Fig. \ref{fig:fig1_system}.

In the data acquisition module, we use the Microsoft Kinect camera to record the subjects’ motion. Microsoft Kinect capture both the texture and depth information of the scene and provide real-time human skeletal joint data [zhengyou zhang]. Based on the skeletal data estimated by Microsoft Kinect SDK, the data processing module remove the outlier data caused by loss tracking with the joint Gaussian and uniform distribution model introduced in Section 3. It further extracts the kinematic parameters from the skeletal motion data. The performance evaluation module does real-time motion analysis, including repetition counting, most moving joints, reachable range, and moving time, which can be further used to measure the flexibilities, endurance, strength, and balance of the subjects. The feedback module can illustrate the evaluation results in the user interface on the screen and also provide both the system automatically generated suggestions and doctors’ suggestion to the subject. Moreover, the system store the subjects’ motion data in the database for the further reference in the future.

\section{Kinect Data Accuracy Evaluation}

\subsection{Data accuracy evaluation}

\begin{figure}[t]
	\begin{center}
		\includegraphics[width=0.9\linewidth]{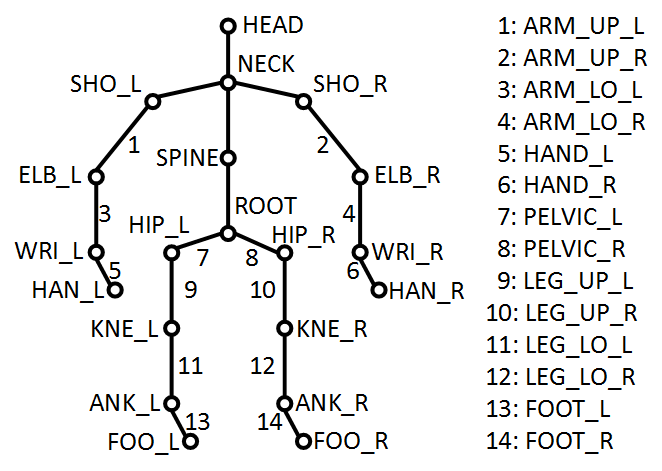}
	\end{center}
	\caption{Diagram of the 20-joint skeleton with labeled joint and bone segments that are used in the analysis (L-left, R-right, UP-upper, LO-lower).}
	\label{fig:fig2_skeleton}
\end{figure}

In order to quantify the fidelity of the motion data captured by Microsoft Kinect, we evaluated its pose tracking with respect to the motion capture system \cite{wang2015evaluation}. In this study, we evaluate both the first and second generations of Microsoft Kinect which are called Kinect 1 and Kinect 2, respectively, in the following of this section. The motion capture system we use in this evaluation is the PhaseSpace optical motion capture system which can provide high precision motion capture data at 120 Hz. The Kinect 1 and Kinect 2 capture the motion data at 20 Hz. During the evaluation, all these three systems capture the human’s motion simultaneously. The motion data from these three devices are temporally synchronized using Network Time Protocol (NTP). The skeletal joint data from Kinect 1 and Kinect 2 are extracted based on the latest SDK of them. The skeletal joint data from motion capture system are obtained by the Recap2 software. We picked 20 common joints and 14 common segments among the three systems. The joints and segments picked in the evaluation are demonstrated in Fig. \ref{fig:fig2_skeleton}.

During the evaluation, we perform 12 different exercises that six of them are in the sitting pose which are designed by the physical therapist for the elderly people with physical problems. The other six are in the standing pose which is the most common motion that people may perform in their daily live. Each exercise are performed by the subjects for five times and captured in three different view angles, including the frontal view, right side view with 30 degree and 60 degree. The degree here means the angle between the subjects’ frontal direction and the optical axis of the Kinect cameras. 

For the joint position, we evaluate the offset and variance of the joint positions obtained by Kinect 1 and Kinect 2 with respect to the corresponding joint position captured by motion capture system. All the joint positions from these three systems are transformed into a unified coordinate system by the camera calibration performed before the motion capture.

Tables \ref{tab:tab1_joint_pos_sitting} and \ref{tab:tab2_joint_pos_standing} shows the average joint offset and variance of the offset in the sitting and standing exercise, respectively.

\begin{table*}
	\begin{center}
		\caption{Joint position offsets in sitting exercises.}
		\label{tab:tab1_joint_pos_sitting}%
		\begin{tabular}{|l|c|c|c|c|c|c||c|c|c|c|c|c|}
			\hline
			\multirow{3}{*}{} & \multicolumn{6}{c||}{Kinect~1}                  & \multicolumn{6}{c|}{Kinect~2} \\
			\cline{2-13}
			& \multicolumn{3}{c|}{Mean (mm)} & \multicolumn{3}{c||}{SD (mm)} & \multicolumn{3}{c|}{Mean (mm)} & \multicolumn{3}{c|}{SD (mm)} \\
			\cline{2-13}
			& 0$^{\circ}$     & 30$^{\circ}$    & 60$^{\circ}$    & 0$^{\circ}$     & 30$^{\circ}$    & 60$^{\circ}$    & 0$^{\circ}$     & 30$^{\circ}$    & 60$^{\circ}$    & 0$^{\circ}$     & 30$^{\circ}$    & 60$^{\circ}$ \\
			\hline
			ROOT  & 256   & 262   & 263   & 25    & 20    & 25    & 100   & 102   & 106   & 17    & 18    & 16 \\
			SPINE & 91    & 97    & 100   & 24    & 20    & 21    & 110   & 117   & 126   & 13    & 14    & 11 \\
			NECK  & 79    & 65    & 62    & 25    & 21    & 23    & 84    & 78    & 73    & 14    & 16    & 14 \\
			HEAD  & 74    & 70    & 67    & 26    & 21    & 21    & 50    & 51    & 50    & 13    & 15    & 12 \\
			SHO\_L & 90    & 89    & 97    & 26    & 24    & 33    & 76    & 76    & 82    & 16    & 24    & 29 \\
			ELB\_L & 81    & 86    & 98    & 27    & 29    & 35    & 87    & 103   & 114   & 17    & 28    & 25 \\
			WRI\_L & 76    & 90    & 118   & 33    & 48    & 55    & 59    & 84    & 115   & 25    & 53    & 44 \\
			HAN\_L & 85    & 106   & 134   & 41    & 60    & 68    & 64    & 95    & 125   & 31    & 60    & 53 \\
			SHO\_R & 78    & 74    & 69    & 28    & 24    & 23    & 80    & 83    & 78    & 17    & 20    & 17 \\
			ELB\_R & 95    & 93    & 89    & 28    & 38    & 34    & 88    & 77    & 70    & 21    & 25    & 19 \\
			WRI\_R & 64    & 93    & 110   & 30    & 53    & 65    & 61    & 64    & 71    & 25    & 28    & 24 \\
			HAN\_R & 83    & 113   & 130   & 35    & 65    & 80    & 74    & 71    & 75    & 24    & 28    & 26 \\
			HIP\_L & 188   & 200   & 215   & 25    & 20    & 22    & 115   & 122   & 139   & 16    & 17    & 15 \\
			KNE\_L & 96    & 95    & 93    & 24    & 24    & 27    & 76    & 95    & 114   & 16    & 18    & 25 \\
			ANK\_L & 54    & 77    & 81    & 17    & 25    & 32    & 93    & 103   & 113   & 16    & 18    & 29 \\
			FOO\_L & 66    & 74    & 86    & 19    & 26    & 38    & 93    & 105   & 119   & 18    & 25    & 39 \\
			HIP\_R & 207   & 210   & 211   & 25    & 21    & 24    & 133   & 128   & 132   & 15    & 18    & 16 \\
			KNE\_R & 123   & 118   & 128   & 23    & 23    & 34    & 120   & 117   & 119   & 15    & 17    & 24 \\
			ANK\_R & 67    & 75    & 91    & 17    & 21    & 27    & 112   & 122   & 132   & 15    & 19    & 27 \\
			FOO\_R & 67    & 74    & 84    & 17    & 20    & 26    & 126   & 134   & 139   & 19    & 20    & 28 \\
			\hline
		\end{tabular}%
	\end{center}
\end{table*}

\begin{table*}
	\begin{center}
		\caption{Joint position offsets in standing exercises.}
		\label{tab:tab2_joint_pos_standing}%
		\begin{tabular}{|l|c|c|c|c|c|c||c|c|c|c|c|c|}
			\hline
			\multirow{3}{*}{} & \multicolumn{6}{c||}{Kinect~1}                  & \multicolumn{6}{c|}{Kinect~2} \\
			\cline{2-13}
			& \multicolumn{3}{c|}{Mean (mm)} & \multicolumn{3}{c||}{SD (mm)} & \multicolumn{3}{c|}{Mean (mm)} & \multicolumn{3}{c|}{SD (mm)} \\
			\cline{2-13}
			& 0$^{\circ}$     & 30$^{\circ}$    & 60$^{\circ}$    & 0$^{\circ}$     & 30$^{\circ}$    & 60$^{\circ}$    & 0$^{\circ}$     & 30$^{\circ}$    & 60$^{\circ}$    & 0$^{\circ}$     & 30$^{\circ}$    & 60$^{\circ}$ \\
			\hline
			ROOT & 245   & 256   & 267   & 23    & 25    & 25    & 76    & 81    & 93    & 24    & 19    & 18\\
			SPINE$*$ & 79    & 89    & 102   & 22    & 22    & 24    & 112   & 129   & 144   & 17    & 16    & 17 \\
			NECK  & 82    & 91    & 102   & 23    & 24    & 23    & 113   & 129   & 143   & 18    & 16    & 16 \\
			HEAD  & 89    & 84    & 87    & 30    & 29    & 26    & 79    & 82    & 90    & 26    & 22    & 20 \\
			SHO\_L & 76    & 76    & 80    & 33    & 36    & 38    & 68    & 72    & 78    & 29    & 31    & 31 \\
			ELB\_L & 98    & 112   & 134   & 46    & 52    & 66    & 112   & 137   & 159   & 37    & 41    & 57 \\
			WRI\_L & 85    & 93    & 110   & 56    & 57    & 71    & 66    & 84    & 111   & 47    & 52    & 73 \\
			HAN\_L & 85    & 96    & 114   & 62    & 65    & 80    & 77    & 94    & 121   & 49    & 55    & 80 \\
			SHO\_R & 76    & 76    & 74    & 30    & 30    & 26    & 96    & 106   & 109   & 26    & 24    & 22 \\
			ELB\_R & 98    & 89    & 80    & 38    & 39    & 34    & 96    & 91    & 80    & 33    & 32    & 31 \\
			WRI\_R & 80    & 82    & 80    & 43    & 43    & 38    & 78    & 88    & 87    & 33    & 32    & 31 \\
			HAN\_R & 85    & 83    & 77    & 46    & 49    & 44    & 81    & 79    & 74    & 33    & 34    & 32 \\
			HIP\_L & 186   & 205   & 228   & 28    & 27    & 27    & 106   & 125   & 150   & 23    & 19    & 21 \\
			KNE\_L & 101   & 111   & 124   & 33    & 38    & 52    & 107   & 129   & 148   & 26    & 34    & 51 \\
			ANK\_L & 119   & 144   & 135   & 43    & 61    & 58    & 135   & 168   & 174   & 33    & 51    & 58 \\
			FOO\_L & 115   & 135   & 122   & 42    & 60    & 59    & 130   & 170   & 180   & 33    & 56    & 63 \\
			HIP\_R & 188   & 195   & 207   & 25    & 24    & 23    & 103   & 104   & 109   & 25    & 17    & 17 \\
			KNE\_R & 105   & 102   & 101   & 31    & 29    & 26    & 115   & 119   & 118   & 26    & 25    & 27 \\
			ANK\_R & 111   & 113   & 108   & 38    & 39    & 28    & 146   & 158   & 157   & 32    & 33    & 35 \\
			FOO\_R & 97    & 93    & 86    & 36    & 37    & 29    & 146   & 157   & 150   & 32    & 33    & 38 \\
			\hline
		\end{tabular}%
	\end{center}
\end{table*}

We also evaluate the segment length based on the motion data. Tables \ref{tab:tab3_bone_length_sitting} and \ref{tab:tab4_bone_length_standing} demonstrate the average segment length error and variance of the segment length in the sitting and standing exercises.

\begin{table*}[!htbp] 
	\begin{center}
		\caption{Mean and SD of the bone length difference, sitting pose}
		\label{tab:tab3_bone_length_sitting}%
		\begin{tabular}{|l|c|c|c|c|c|c||c|c|c|c|c|c|}
			\hline
			\multirow{3}{*}{} & \multicolumn{6}{c||}{Kinect~1}                  & \multicolumn{6}{c|}{Kinect~2} \\
			\cline{2-13}
			& \multicolumn{3}{c|}{Mean (mm)} & \multicolumn{3}{c||}{SD (mm)} & \multicolumn{3}{c|}{Mean (mm)} & \multicolumn{3}{c|}{SD (mm)} \\
			\cline{2-13}
			& 0$^{\circ}$     & 30$^{\circ}$    & 60$^{\circ}$    & 0$^{\circ}$     & 30$^{\circ}$    & 60$^{\circ}$    & 0$^{\circ}$     & 30$^{\circ}$    & 60$^{\circ}$    & 0$^{\circ}$     & 30$^{\circ}$    & 60$^{\circ}$ \\
			\hline
			ARM\_UP\_L  & -76   & -74   & -81   & 17    & 18    & 21    & -67   & -62   & -68   & 14    & 17    & 20 \\
			ARM\_LO\_L & -46   & -40   & -32   & 17    & 22    & 27    & -39   & -30   & -24   & 14    & 19    & 23 \\
			HAND\_L & -3    & -4    & -3    & 22    & 19    & 20    & -20   & -17   & -17   & 20    & 24    & 25 \\
			ARM\_UP\_R & -68   & -70   & -69   & 17    & 17    & 18    & -67   & -67   & -69   & 15    & 14    & 13 \\
			ARM\_LO\_R & -37   & -30   & -29   & 17    & 20    & 21    & -35   & -28   & -30   & 13    & 15    & 16 \\
			HAND\_R & 2     & 3     & 3     & 21    & 24    & 24    & -15   & -13   & -9    & 20    & 19    & 17 \\
			PELVIC\_L & -28   & -28   & -36   & 6     & 5     & 8     & -54   & -62   & -75   & 3     & 3     & 4 \\
			LEG\_UP\_L & 49    & 38    & 38    & 30    & 32    & 34    & 9     & 5     & 5     & 16    & 18    & 18 \\
			LEG\_LO\_L & -90   & -100  & -81   & 23    & 28    & 31    & -80   & -79   & -75   & 17    & 18    & 21 \\
			FOOT\_L & -13   & -4    & 1     & 8     & 10    & 13    & 36    & 38    & 32    & 12    & 14    & 19 \\
			PELVIC\_R & -29   & -33   & -41   & 6     & 5     & 7     & -52   & -47   & -49   & 3     & 3     & 4 \\
			LEG\_UP\_R & 44    & 46    & 53    & 31    & 28    & 31    & 1     & 9     & 18    & 16    & 18    & 18 \\
			LEG\_LO\_R & -86   & -88   & -95   & 25    & 26    & 31    & -76   & -82   & -74   & 17    & 19    & 23 \\
			FOOT\_R & -8    & -2    & 8     & 7     & 9     & 11    & 41    & 44    & 40    & 12    & 11    & 12 \\
			\hline
		\end{tabular}%
	\end{center}
\end{table*}%

\begin{table*}[!htbp] 
	\begin{center}
		\caption{Mean and SD of the bone length difference, standing pose}
		\label{tab:tab4_bone_length_standing}%
		\begin{tabular}{|l|c|c|c|c|c|c||c|c|c|c|c|c|}
			\hline
			\multirow{3}{*}{} & \multicolumn{6}{c||}{Kinect~1}                  & \multicolumn{6}{c|}{Kinect~2} \\
			\cline{2-13}
			& \multicolumn{3}{c|}{Mean (mm)} & \multicolumn{3}{c||}{SD (mm)} & \multicolumn{3}{c|}{Mean (mm)} & \multicolumn{3}{c|}{SD (mm)} \\
			\cline{2-13}
			& 0$^{\circ}$     & 30$^{\circ}$    & 60$^{\circ}$    & 0$^{\circ}$     & 30$^{\circ}$    & 60$^{\circ}$    & 0$^{\circ}$     & 30$^{\circ}$    & 60$^{\circ}$    & 0$^{\circ}$     & 30$^{\circ}$    & 60$^{\circ}$ \\
			\hline
			ARM\_UP\_L & -60   & -62   & -65   & 19    & 22    & 23    & -54   & -54   & -51   & 15    & 18    & 20 \\
			ARM\_LO\_L & -30   & -31   & -34   & 16    & 18    & 20    & -34   & -33   & -32   & 17    & 17    & 18 \\
			HAND\_L & -7    & -5    & 0     & 21    & 21    & 23    & -32   & -32   & -27   & 20    & 21    & 23 \\
			ARM\_UP\_R & -53   & -58   & -59   & 21    & 19    & 18    & -59   & -66   & -66   & 17    & 14    & 14 \\
			ARM\_LO\_R & -22   & -22   & -22   & 17    & 18    & 20    & -30   & -27   & -25   & 18    & 18    & 17 \\
			HAND\_R & -1    & 4     & 6     & 23    & 21    & 19    & -21   & -16   & -14   & 19    & 18    & 18 \\
			PELVIC\_L & -27   & -30   & -37   & 9     & 7     & 8     & -52   & -63   & -75   & 5     & 5     & 5 \\
			LEG\_UP\_L & 113   & 119   & 137   & 28    & 33    & 39    & -13   & -10   & 4     & 19    & 21    & 25 \\
			LEG\_LO\_L & -64   & -66   & -65   & 29    & 29    & 31    & -79   & -85   & -86   & 21    & 23    & 25 \\
			FOOT\_L & -6    & -1    & 2     & 9     & 12    & 12    & 43    & 42    & 34    & 9     & 12    & 14 \\
			PELVIC\_R & -28   & -34   & -41   & 7     & 6     & 8     & -47   & -44   & -44   & 7     & 4     & 5 \\
			LEG\_UP\_R & 103   & 115   & 128   & 29    & 27    & 29    & -25   & -18   & -6    & 18    & 18    & 20 \\
			LEG\_LO\_R & -61   & -57   & -58   & 26    & 25    & 25    & -79   & -79   & -77   & 20    & 20    & 20 \\
			FOOT\_R & -2    & 1     & 4     & 8     & 9     & 10    & 48    & 45    & 32    & 9     & 11    & 16 \\
			\hline
		\end{tabular}%
	\end{center}
\end{table*}%

The results shows that overall Microsoft Kinect 2 generates more stable skeletal joint tracking with less offset and variance as compared to Microsoft Kinect 1. The variance of the joint position and segment length are both increasing with the view angle increasing. Some joints usually have larger offset and variance than the other joints, e.g. the hip and knee joints in Kinect 1, the ankle and foot joints in Kinect 2.

\subsection{Outlier data removal}

The joint position offsets and variance in general depend on various sources of error, such as systematic errors, noise from depth quantization, occlusions, and ambiguity, etc. Compared to the multi-view motion capture system, the Kinect camera captures the motion data from a single view-point \cite{Zhang_2012}. The occlusions and ambiguity from a single view-point will cause loss of tracking. Although the Kinect can infer the joint position based on the related body part \cite{Shotton_2011}, it is not reasonable to perform motion data analysis based on the inferred motion data due to tracking loss. Therefore, we analyze the error distribution to discriminate between the random errors and the errors due to tracking loss.

Generally, the random systematic errors follow the Gaussian distribution while the errors due to tracking loss can be treated as outliers belonging to a uniform distribution. Therefore, we use a mixture model of a Gaussian distribution and a uniform distribution to approximate the distribution of the joint position offsets, as defined in equation \ref{equ:mixture_model}. 

\begin{equation}
p(\theta) = \rho\times N(\mu,\sigma)+(1-\rho)\times U(x_1, x_2).
\label{equ:mixture_model}
\end{equation}

\begin{figure}[t] 
	\begin{center}
		\begin{tabular}{cc}
			\includegraphics[trim=0 0 0 20, clip, width=4cm]{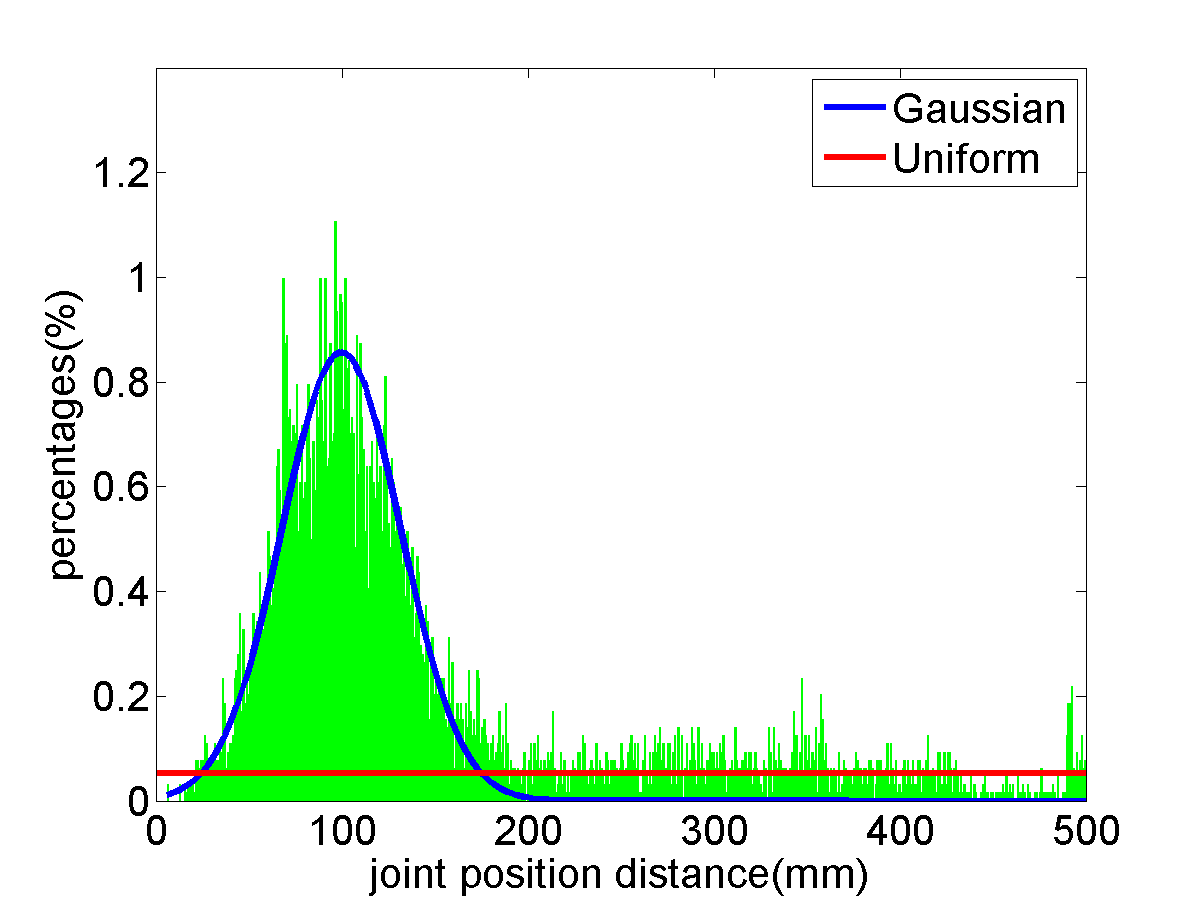}&
			\includegraphics[trim=0 0 0 20, clip, width=4cm]{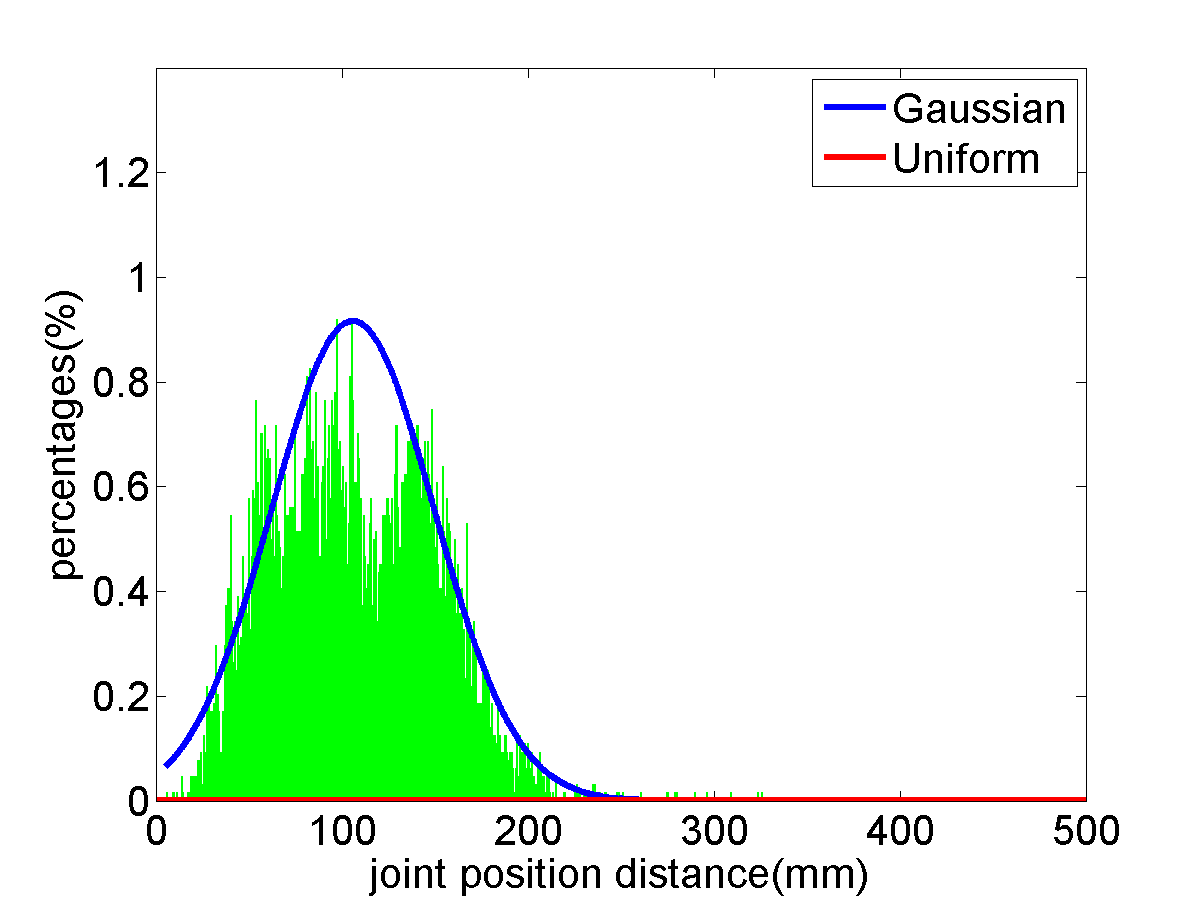}\\
			Kinect~1  & Kinect~2   \\
		\end{tabular}
		\caption{Mixture model fitting into the distribution of joint position offsets for the right elbow (viewpoint: 60$^{\circ}$).}
		\label{fig:fig3_joint_distribution_fitting}
	\end{center}
\end{figure}

In equation \ref{equ:mixture_model}, $p$ denote the mixture distribution of joint position, $\theta$, $N(\mu,\sigma)$ and $U(x_1, x_2)$ denote the Gaussian and uniform distribution, respectively, $\rho$ denotes the weighting parameter. Fig. \ref{fig:fig3_joint_distribution_fitting} demonstrates the distribution fitting results for the right elbow in the exercise. The results show the mixture model of the Gaussian and uniform distributions overlaid on the data histograms.

By applying the mixture model and maximal likelihood estimation, we can remove the outlier data which satisfy the uniform distribution and make cleaner data for the further analysis. Table I demonstrates the joint offset of both standing and sitting exercises without the outlier data.

\begin{table}[!htbp] 
	\begin{center}
		\caption{Average joint position offsets without outliers. }
		\begin{tabular}{|l|l|c|c|c||c|c|c|}
			\hline
			\multirow{2}{*}{} & \multirow{2}{*}{} & \multicolumn{3}{c|}{Mean (\%)} & \multicolumn{3}{c|}{SD (\%)}  \\
			\cline{3-8}
			& & 0$^{\circ}$     & 30$^{\circ}$    & 60$^{\circ}$    & 0$^{\circ}$     & 30$^{\circ}$    & 60$^{\circ}$  \\
			\hline
			\multirow{2}{*}{Kinect 1} & Sitting & 5     & 8     & 6     & 35    & 38    & 35    \\
		        	& Standing & 8     & 7     & 8     & 39    & 36    & 33   \\
			\hline
			\multirow{2}{*}{Kinect 2} & Sitting & 2 & 5     & 3     & 23    & 36    & 23  \\
			         & Standing & 8 & 5     & 6     & 49    & 46    & 43  \\
			\hline
		\end{tabular}%
		\label{tab:tab5_joint_pos_outlier_gain}%
	\end{center}
\end{table}%


\section{Kinematic parameter estimation}
The kinematic parameters, such as the segment length, joint quaternion, and positions are widely used in human motion analysis and human robotic interaction. For the high fidelity motion data, the kinematic parameters can be directly estimated from the skeletal joint position. However, generating the kinematic parameters from the noisy motion data captured by Kinect will propagate the noise into the following processing and analysis steps. In order to reduce the noise in the derived kinematic parameters from the Kinect noise motion data, we propose a kinematic filtering approach in this section.

\subsection{Kinematic model}
In motion analysis, the human’s skeleton can be represented as a series joints which are connected by bones. Therefore, the each joint position can be derived by its parent joint position, the corresponding segment orientation and length. The human’s skeleton can thus be represented as a kinematic chain. In this paper, we treat the root joint as the root of each kinematic chain. Based on the skeletal structure shown in Fig. \ref{fig:fig2_skeleton}, the SPINE, NECK, SHO\_L, SHO\_R, HIP\_L, and HIP\_R are modeled as the 3-DoF joints and their parent joints are ROOT, SPINE, NECK, NECK, ROOT, ROOT, respectively; The ELB\_L, ELB\_R, KNE\_L, KNE\_R are treated as the 1-DoF joints whose parent joints are SHO\_L, SHO\_R, HIP\_L, HIP\_R, respectively; The WRI\_L, WRI\_R, ANK\_L, and ANK\_R are the 2-DoF joints that their parent joints are ELB\_L, ELB\_R, KNE\_L, KNE\_R, respectively. For each joint, its position can be derived by its parent joint position, $\mathbf{T}_p$, the segment length of its corresponding segment, $l_c$, the relative rotation matrix of its corresponding segment, $\mathbf{R}_l$, with respect to its parent segment, the rotation matrix of its parent segment, $\mathbf{R}_p$, as the following function $\mathbf{F}$:
\begin{equation}
\begin{split}
[\mathbf{T}_c, 1]^T &= F(\mathbf{T}_p, l_c, \mathbf{R}_l, \mathbf{R}_p) \\
& = \begin{bmatrix} \mathbf{R}_l\times\mathbf{R}_p & \mathbf{T}_p \\ 0 & 1 \end{bmatrix} \times \left[ \begin{array}{c} 0\\ 0\\ l_c \\ 1 \end{array} \right]
\end{split}
\label{equ:joint_transform}
\end{equation}

\subsection{Kinematic filtering}

With the kinematic model presented in the last section, we propose a kinematic filtering algorithm based on Kalman filtering to generate the kinematic parameter from the Kinect motion data \cite{wang2015unsupervised}. The input state vector is constituted by the root joint position, the segment length, and the segment rotation quaternion. The observation vector is constituted by the rest joint positions. The state transition is modeled by a random-walk process. The observation measuring function is a set of the kinematic model function as defined by equation \ref{equ:joint_transform}. Since the state transition and the observation measuring functions are nonlinear, we apply the Unscented Kalman Filter (UKF) \cite{wan2000unscented} for the filtering. Since the segment lengths for a certain person are constants, we propose a four-pass UKF to meet with this constraints. In the first forward pass, the UKF is defined by the following functions

\begin{equation}
\mathbf{x}(t) = \mathbf{x}(t-1)+\boldsymbol{\delta}(t)
\label{equ:ukf_state_forward}
\end{equation}

\begin{equation}
\mathbf{y}(t) = \mathbf{F}(\mathbf{x}(t))+\boldsymbol{\varepsilon}(t)
\label{equ:ukf_observation}
\end{equation}

In equations \ref{equ:ukf_state_forward} and \ref{equ:ukf_observation}, the $\mathbf{x}(t)$ and $\mathbf{x}(t+1)$ denote the state vector at time $t$ and $t+1$, respectively, $\boldsymbol{\delta}$ denote the random term in the random walk process, $\mathbf{y}$ denote the observation vector, $\boldsymbol{\varepsilon}$ denotes the observation noise. In the first backward pass, the observation measuring function is the same as equation \ref{equ:ukf_observation}. The state transition function is defined as 
\begin{equation}
\mathbf{x}(t) = \mathbf{x}(t+1)+\boldsymbol{\delta}(t)
\label{equ:ukf_state_backward}
\end{equation}

After the first two pass, the length of each segment are assigned as a constant in the following forward and backward passes. The covariance and the random terms in the random walk process of the segment lengths are all set to zeros.

\begin{figure}[t]
	\begin{center}
		\includegraphics[width=0.9\linewidth]{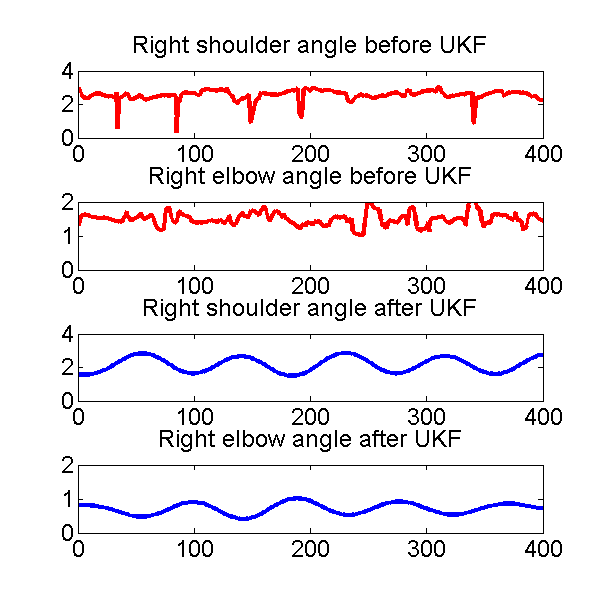}
	\caption{Joint angles of the shoulder and elbow calculated by the input raw motion data and the output kinematic parameters, respectively, horizontal axes represent the frame count, vertical axes represent the joint angle.}
	\label{fig:fig4_joint_angle_filtering}
	\end{center}
\end{figure}

Fig. \ref{fig:fig4_joint_angle_filtering} demonstrates the filtering results of UKF for the motion data from Kinect 1. The segment quaternion curves after filtering are much smoother than the input noisy data and also demonstrate strong periodic pattern which is in accordance to the input motion. Therefore, the UKF based on the kinematic model is a practical solution to extract kinematic parameters.


\section{Conclusions}

This paper summarize the recent progress in the motion analysis for physical therapy with the affordable and accessible device like Microsoft Kinect. Based on preliminary studies, our remote health coaching system receive positive feedback from the users for its accessibility and interaction. Since the motion data captured by Microsoft Kinect is noisy, we further studied the data fidelity of each joint position and segment length. We also propose a mixture model to distinguish the outlier data caused by loss of tracking. It can helps to reduce the noise in the further motion data analysis. Finally, the kinematic filtering based on UKF provide a solution to extract smooth kinematic parameters from the noisy motion data.


{\small
\bibliographystyle{ieee}
\bibliography{egbib}
}

\end{document}